\documentclass[letterpaper]{article} 
\usepackage{xxxx}  
\usepackage{times}  
\usepackage{helvet}  
\usepackage{courier}  
\usepackage[hyphens]{url}  
\usepackage{graphicx} 
\usepackage{natbib}  
\usepackage{caption} 
\usepackage{algorithm}
\usepackage{algorithmic}
\usepackage{amsmath}
\usepackage{amssymb}
\usepackage{threeparttable}
\usepackage{booktabs}
\usepackage{multirow}

\usepackage{newfloat}
\usepackage{listings}
\DeclareCaptionStyle{ruled}{labelfont=normalfont,labelsep=colon,strut=off} 
\floatstyle{ruled}
\newfloat{listing}{tb}{lst}{}
\floatname{listing}{Listing}
\nocopyright

\title{DrivingForward: Feed-forward 3D Gaussian Splatting for Driving Scene Reconstruction from Flexible Surround-view Input}
\author {
    Qijian Tian\textsuperscript{\rm 1},
    Xin Tan\textsuperscript{\rm 2}\thanks{This project is led by Xin Tan.},
    Yuan Xie\textsuperscript{\rm 2},
    Lizhuang Ma\textsuperscript{\rm 1 2 3}
}
\affiliations {
    \textsuperscript{\rm 1}Shanghai Jiao Tong University, Shanghai, China\\
    \textsuperscript{\rm 2}East China Normal University, Shanghai, China\\
    \textsuperscript{\rm 3}MoE Key Lab of Artificial Intelligence, Shanghai Jiao Tong University, Shanghai, China\\
    tianqijian@sjtu.edu.cn, xtan@cs.ecnu.edu.cn, xieyuan8589@foxmail.com, ma-lz@cs.sjtu.edu.cn\\
}

\usepackage{bibentry}

\begin{document}

\maketitle

\begin{abstract}
We propose DrivingForward, a feed-forward Gaussian Splatting model that reconstructs driving scenes from flexible surround-view input. Driving scene images from vehicle-mounted cameras are typically sparse, with limited overlap, and the movement of the vehicle further complicates the acquisition of camera extrinsics. To tackle these challenges and achieve real-time reconstruction, we jointly train a pose network, a depth network, and a Gaussian network to predict the Gaussian primitives that represent the driving scenes. The pose network and depth network determine the position of the Gaussian primitives in a self-supervised manner, without using depth ground truth and camera extrinsics during training. The Gaussian network independently predicts primitive parameters from each input image, including covariance, opacity, and spherical harmonics coefficients. At the inference stage, our model can achieve feed-forward reconstruction from flexible multi-frame surround-view input. Experiments on the nuScenes dataset show that our model outperforms existing state-of-the-art feed-forward and scene-optimized reconstruction methods in terms of reconstruction.\par
\end{abstract}

%
\begin{links}
    \link{Code}{https://github.com/fangzhou2000/DrivingForward}
\end{links}

\section{Introduction}
3D scene reconstruction is critical for understanding driving scenes. Modern self-driving assistance cars are usually equipped with several cameras to capture surrounding scenes. Real-time reconstruction of driving scenes from sparse vehicle-mounted cameras contributes to various downstream tasks in autonomous driving, including online mapping~\cite{online_mapping_hdmapnet}, BEV perception~\cite{bev_bevformer, bev_bevfusion1, bev_bevfusion2}, scene understanding~\cite{scene_fastlgs, scene_unitomulti, scene_positive-negative, scene_imageunderstands, scene_cotr} and 3D detection~\cite{3d_detection_futr3d, 3d_detection_objectfusion}. However, the real-time computing required by downstream tasks and the sparse surrounding views challenge the driving scene reconstruction.\par 
Neural Radiance Fields (NeRF)~\cite{nerf} and 3D Gaussian Splatting (3DGS)~\cite{3dgs} have significantly progressed the development of 3D scene reconstruction. DrivingGaussian~\cite{drivinggaussian}, StreetGaussian~\cite{streetgaussian}, S$^3$Gaussian~\cite{s3gaussian}, and AutoSplat~\cite{, autosplat} further explore the reconstruction of driving scenes. While these methods demonstrate strong capability in novel view synthesis, they are scene-optimized methods that require dozens of images and expensive computing time to reconstruct just one scene. These offline reconstruction methods are unsuitable for real-time downstream tasks in autonomous driving, thereby limiting their practicality.\par
\begin{figure}[t]
    \centering
    \includegraphics[width=0.95\linewidth]{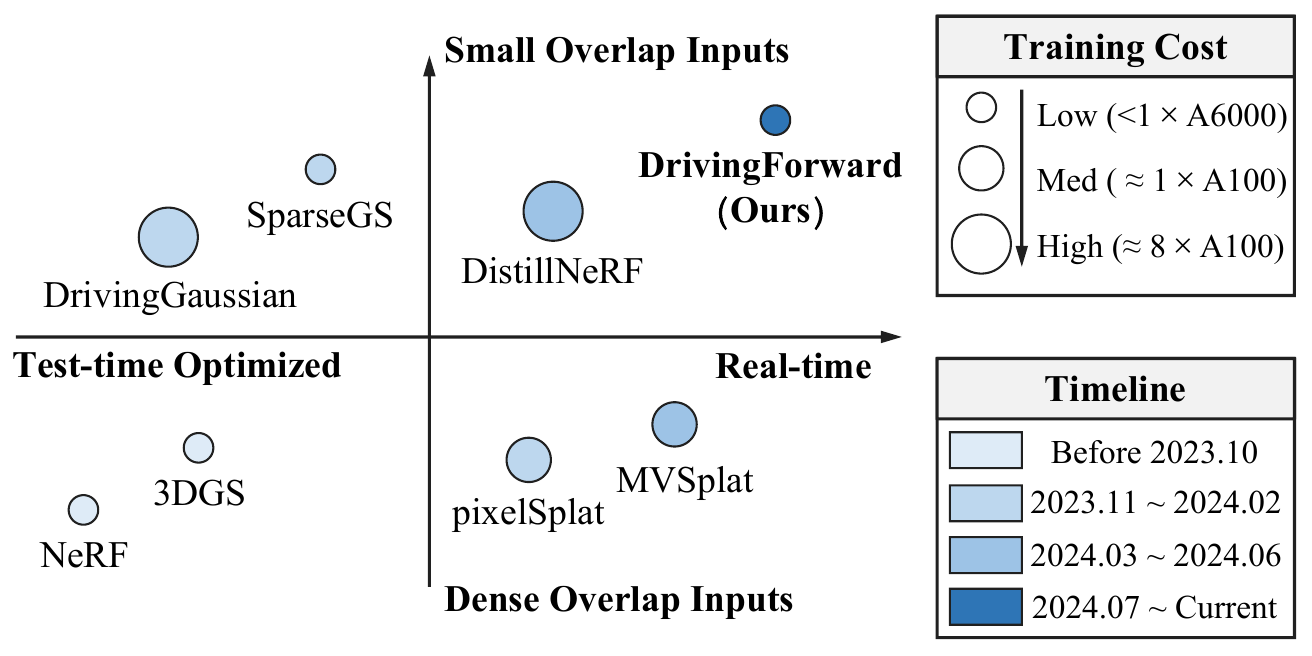}
    \caption{Comparison of our DrivingForward with the latest related works. We achieve real-time reconstruction from small overlap inputs with fewer computing resources.}
    \label{fig: intro}
\end{figure}
Our goal is to achieve online, generalizable driving scene reconstruction from sparse surrounding views. Several attempts, such as pixelSplat~\cite{pixelsplat} and MVSplat~\cite{mvsplat}, have explored the generalizable reconstruction. They learn powerful priors from large-scale datasets during training and achieve fast 3D reconstruction from sparse input views through a feed-forward inference. Unfortunately, these methods are difficult to apply in driving scenes. Since the number of vehicle-mounted cameras is limited (usually 6 cameras), the overlap of adjacent views is extremely small (as low as 10\%). While these existing methods often require densely overlapping (usually over 60\%) input images. Additionally, acquiring camera extrinsics for each view at various timesteps in driving scenes is costly. These methods depend on such data during training, limiting their practical applicability. A recent NeRF-based work DistillNeRF~\cite{distillnerf} attempts to develop a generalizable 3D representation for driving scenes. However, it gains a suboptimal performance and relies on LiDAR to train numerous NeRF models for distillation, which is extremely computationally expensive. Besides, previous feed-forward methods typically have a fixed mode of input views, either using stereo images (e.g., MVSplat, pixelSplat) or single-frame images of surrounding views (e.g., DistillNeRF). However, as the vehicle moves forward and captures surround-view images frame-by-frame, we aim to support flexible multi-frame inputs for reconstruction, such as predicting the next frame's views from single-frame surrounding views or synthesizing intermediate frame surrounding views from two interval frames.\par
In summary, online and generalizable reconstruction of driving scenes face challenges including real-time processing, sparse surrounding views with minimal overlap, and variable numbers of input frames.\par
To this end, we introduce DrivingForward, a novel feed-forward Gaussian Splatting model that enables real-time reconstruction of driving scenes from flexible sparse surround-view images. We train a generalizable model and achieve real-time reconstruction via a feed-forward inference. In driving scenes, the minimal overlap between sparse cameras limits the direct use of geometric relationships from multi-views. Consequently, we predict Gaussian primitives from each input image individually and aggregate them to represent the 3D driving scenes. However, reconstruction from a single image is inherently ill-posed due to the principle of scale ambiguity~\cite{pixelsplat}, which may lead to inconsistent scales across multi-views. To address this issue, inspired by surround-view depth estimation~\cite{vfdepth, fsm}, we propose scale-aware localization for Gaussian primitives. At the training stage, we input multi-frame surround-view images into a pose network and a depth network. The pose network predicts the camera pose, i.e., extrinsics, and the depth network estimates the dense depth map for each image. These two networks are only supervised by the photometric loss from input images and learn scale information in a self-supervised manner without ground truth depth and camera extrinsics. At the inference stage, the depth network predicts real-scale depth from single-frame images individually, ensuring consistent depth estimation across multi-frame inputs.\par
By unprojecting the consistent depth estimation, we obtain the position of Gaussian primitives. For other Gaussian parameters, we individually predict them from each image through a Gaussian network. The Gaussian network is jointly trained with the pose network and depth network. It takes the depth map and image feature from the depth network as input and outputs the covariance, opacity, and spherical harmonics coefficients of Gaussian primitives. 
Since Gaussian primitives are predicted independently from single-frame images of surrounding views, our method is not constrained by a fixed number of input frames. This allows for flexible multi-frame surround-view input, such as predicting the next frames’ views from the current frame or synthesizing the intermediate frame from two interval frames.\par
Extensive experiments on the nuScenes dataset demonstrate that our DrivingForward outperforms other feed-forward methods on the novel view synthesis under various inputs. It also achieves higher reconstruction quality compared to scene-optimized methods with the same input. A functional comparison of our DrivingForward with the latest related works is present in Figure~\ref{fig: intro}.\par
We summarize our main contributions as follows:
\begin{itemize}
    \item To our knowledge, DrivingForward is the first feed-forward Gaussian Splatting model for surround-view driving scenes. It achieves real-time reconstruction from sparse vehicle-mounted cameras and supports flexible multi-frame inputs of surrounding views.\par
    \item We introduce a scale-aware localization and Gaussian parameters prediction to reconstruct driving scenes. The scale-aware localization learns real scale depth from surrounding views without using ground truth depth and extrinsics. Then we independently predict Gaussian parameters from each image, thereby supporting flexible multi-frame inputs. The full model is trained end-to-end.\par
    \item Comprehensive experiments show that DrivingForward achieves the best performances against both feed-forward methods and scene-optimized methods for driving scene reconstruction.\par
\end{itemize}
\begin{figure*}[t!]
    \centering
    \includegraphics[width=.95\textwidth]{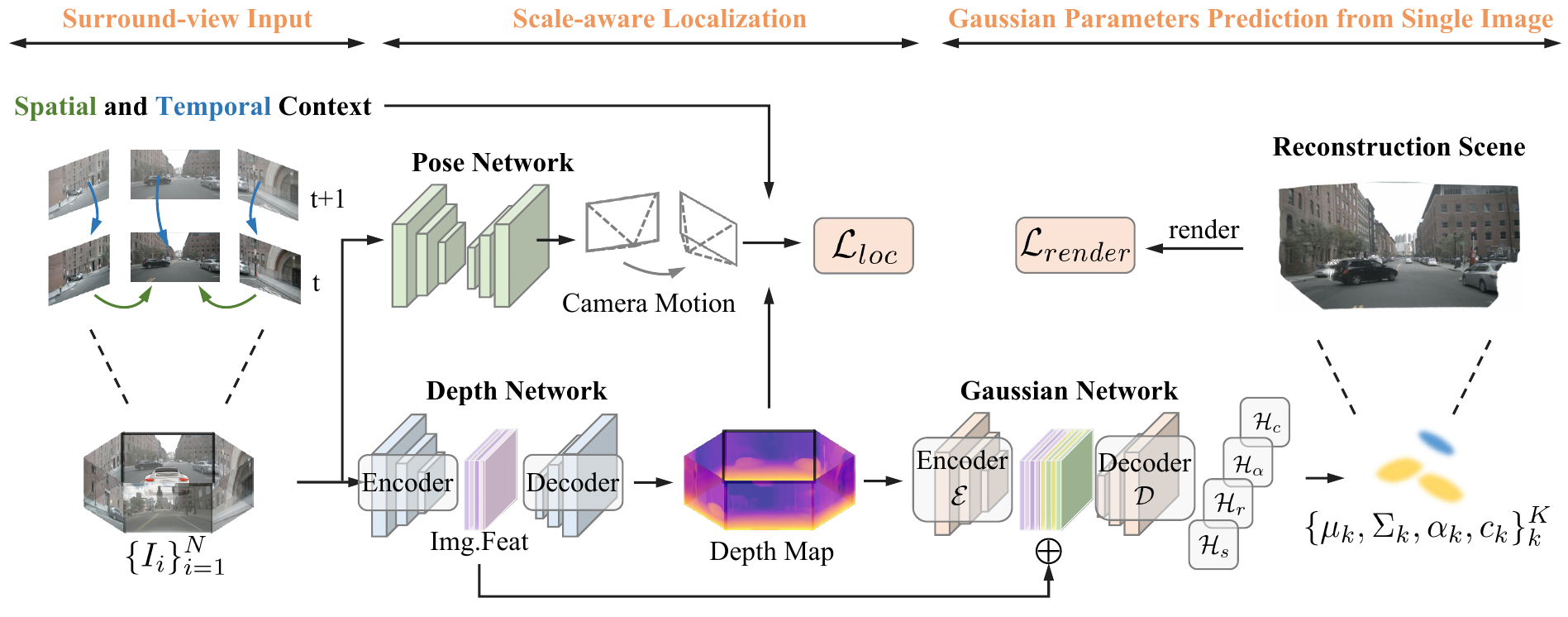}
    \caption{Overview of DrivingForward. Given sparse surround-view input from vehicle-mounted cameras, our model learns scale-aware localization for Gaussian primitives from the small overlap of spatial and temporal context views. A Gaussian network predicts other parameters from each image individually. This feed-forward pipeline enables the real-time reconstruction of driving scenes and the independent prediction from single-frame images supports flexible input modes. At the inference stage, we include only the depth network and the Gaussian network, as shown in the lower part of the figure.}
    \label{fig: main}
\end{figure*}
\section{Related Work}
\subsection{Feed-Forward Reconstruction}
Neural Radiance Fields (NeRF) \cite{nerf} and 3D Gaussian Splatting (3DGS) \cite{3dgs} have significantly advanced 3D scene reconstruction. Some following works~\cite{sparsegs, dngaussian} also explore the reconstruction from sparse views. However, these scene-optimized methods require training on dozens of images for each scene and lack generalizability across different scenes. In contrast, feed-forward reconstruction methods learn powerful priors from large-scale datasets and reconstruct scenes through a feed-forward inference from sparse views, which are significantly faster than scene-optimized methods. NeRF-based methods~\cite{pixelnerf, du, gpnr} pioneer the paradigm of feed-forward reconstruction. Recent 3DGS avoids NeRF’s expensive volume sampling via splat-based rasterization. Therefore, 3DGS-based feed-forward reconstruction methods, such as Splatter Image~\cite{splatter}, GPS-Gaussian~\cite{gps-gaussian}, pixelSplat~\cite{pixelsplat}, and MVSplat~\cite{mvsplat}, outperform the previous NeRF-based methods. However, they fail to apply in driving scenes. Splatter Image focuses on single-object reconstruction, while GPS-Gaussian targets human reconstruction. They are both unsuitable for much larger driving scenes. The pixelSplat involves a two-view encoder to resolve scale ambiguity and MVSplat relies on stereo images to construct cost volume, both requiring densely overlapping images as input. However, the surrounding views of driving scenes have only minimal overlap, which greatly impacts their performance of reconstruction from single-frame surround-view images. In this work, we propose a feed-forward model to reconstruct driving scenes from sparse surrounding views with minimal overlap.\par
\subsection{Driving Scene Reconstruction}
Based on NeRFs or 3DGS, a few methods~\cite{streetsurf, emernerf, drivinggaussian, streetgaussian, autosplat} extend reconstruction specifically for driving scenes. However, most of them focus on accurate 3D or 4D reconstruction for a single scene. These offline reconstruction methods limit the application of real-time downstream tasks in driving scenes. In contrast, our method is designed for real-time reconstruction to support online downstream tasks. A few NeRFs-related methods~\cite{selfocc, unipad} also learn 3D representations for online driving, while they mainly focus on other downstream tasks instead of reconstruction. Another related work DistillNeRF~\cite{distillnerf} proposes a generalizable model for driving scenes and achieves reasonable novel-view synthesis. However, it requires a pre-training stage that involves offline optimization of NeRFs for each scene, which is extremely costly and necessitates 8 A100 GPUs (8 × 80GB). In contrast, our DrivingForward only requires a single GPU with 48GB. DistillNeRF also relies on LiDAR for training, while our method is trained from RGB images alone. 
Despite DitillNeRF requiring more training resources and additional LiDAR data, it achieves only suboptimal performance compared to our DrivingForward under the same setting.\par
\section{Method}
\subsection{Overview}
DrivingForward learns powerful priors from large-scale driving scene datasets during training and achieves real-time driving scene reconstruction in a feed-forward manner from sparse vehicle-mounted cameras at the inference stage.\par
We begin with N sparse camera images $\{I_i\}_{i=1}^N$ as input and aim to predict Gaussian primitives from input view images. The overview framework is illustrated in Figure~\ref{fig: main}. A pose network $\mathcal{P}$ and a deep network $\mathcal{D}$ predict the vehicle motion and estimate the scale-aware depth from the input. We assign each pixel to one Gaussian primitive and the position is located through the estimated depth. Other parameters of the Gaussian primitives are predicted by a Gaussian network $\mathcal{G}$. We unproject the Gaussian primitives from all views into 3D space, render them to the target view in a differentiable way, and jointly train the full model end-to-end. At the inference stage, the depth network and Gaussian network are used for feed-forward reconstruction. Since the scale-aware localization and prediction of other parameters do not depend on other frames, we can flexibly input different numbers of surround-view frames during inference.\par
\subsection{Scale-aware Localization}\label{localization}
The original 3DGS~\cite{3dgs} explicitly models a scene with a set of Gaussian primitives $\{g_k = \{\mu_k, \Sigma_k, \alpha_k, c_k\}_k^K\}$, each of which is parameterized by a position $\mu_k $, a 3D covariance matrix $\Sigma_k$, an opacity $\alpha_k$ and spherical harmonics $c_k$ that determines color. It uses Structure from Motion to initialize the Gaussians' position and optimizes them through splat-based rasterization rendering. \par
In contrast, to achieve feed-forward inference without test-time optimization, we directly predict Gaussian primitives from input images in a pixel-wise manner and assign each pixel to one primitive. In this way, accurately localizing the position of Gaussian primitives is the key to high-quality reconstruction as it determines the center of primitives. However, in driving scenes, the limited overlap between sparse cameras limits geometric relationships from multiple views. This presents challenges for existing methods~\cite{mvsplat, pixelsplat} that depend on multi-views with large overlapping (such as the adjacent frames of the same camera) to get the position. We instead estimate the depth map of a single frame without relying on other frames. To obtain multi-frame consistent depth, we propose a scale-aware localization inspired by self-supervised surround-view depth estimation~\cite{fsm, vfdepth}. It learns scale-aware depth from multi-frame surround views during training and independently predicts the depth of the real scale from different frames of surrounding views during inference, thereby achieving consistent scale-aware Gaussian localization.\par
Specifically, we introduce a pose network $\mathcal{P}$ and a depth network $\mathcal{D}$. At the training stage, we input multi-frame surround view images from sparse vehicle-mounted cameras $\{C_i\}_{i=1}^N$. The pose network predicts the vehicle motion and the depth network estimates the depth map:
\begin{equation}
\begin{aligned}
    \mathcal{P}(I_i^t, I_i^{t'}) &\rightarrow T_i^{t \rightarrow t'}, \\
    \mathcal{D}(I_i^t) &\rightarrow D_i^t,
\end{aligned}
\end{equation}
where $I_i^t$ denotes the image from camera $C_i$ at the timestep $t$, $t' \in \{t+1, t-1\}$, $T_i^{t \rightarrow t'}$ is a project matrix from timestep $t$ to timestep $t'$ that indicates the camera motion and the $D_i^t$ is the depth map of image $I_i^t$.\par 
To learn the camera motion and scale-aware depth map from input images in a self-supervised manner, we apply photometric loss for multi-frame surrounding views. The photometric loss is used to minimize the projecting loss between a target image $I_{trg}$ and a synthesis image $\hat{I}_{trg}$ that is warped from a source image $I_{src}$, and $I_{src}$ is usually obtained from stereo pairs or monocular videos~\cite{monodepth, monodepth2},
\begin{equation}\label{eq: photometric}
    \mathcal{L}_{reproj} = \eta \frac{1 - SSIM(I_{trg}, \hat{I}_{trg})}{2} + (1 - \eta) \lVert I_{trg} - \hat{I}_{trg}\rVert, 
\end{equation}
where SSIM is the structure similarity metric~\cite{ssim}, $\eta$ is 0.15. The warped operation can be depicted as:
\begin{equation}
    \hat{I}_{trg} = I_{src}[K_{src}T^{trg \rightarrow src}D_{trg}K_{trg}^{-1}],
\end{equation}
where $K_{src}$ and $K_{trg}$ are camera intrinsics of $I_{src}$ and $I_{trg}$, $D_{trg}$ is the depth map of $I_{trg}$, and $T^{trg \rightarrow src}$ is the cam-to-cam transformation matrix from $I_{trg}$ to $I_{src}$.\par
In driving scenes, with multi-frame surrounding views, we take different inputs as the source image to compute the photometric loss. First, we use the images from the same camera at different frames, denoted as temporal contexts. Then, we use the images from adjacent cameras at the same frame, denoted as spatial contexts. We also combined the two ways, using images from adjacent cameras at different frames, denoted as spatial-temporal contexts. The key insight is to leverage the small overlap between spatially and temporally neighboring images for matching, which provides scale information and enables learning scale-aware camera motion and depth maps during training.\par
Let $C_i$, $C_j$ be two adjacent cameras and $I_i^t$, $I_j^t$ be their images at timestep $t$. For $i = 1,..., N$:
\begin{equation}
    I_{trg} = I_i^t,
\end{equation}
\begin{equation}
I_{src} = \left\{
\begin{array}{ll}
I_i^{t'}, t' \in \{t+1, t-1\} & \text{for temporal,} \\
I_j^t,                        & \text{for spatial,} \\
I_j^{t'}, t' \in \{t+1, t-1\} & \text{for spatial-temporal,}
\end{array}
\right.
\end{equation}
and 
\begin{equation}
\begin{array}{c}
T^{trg \rightarrow src} = \\ 
\left\{
\begin{array}{ll}
T_i^{t \rightarrow t'}, t' \in \{t+1, t-1\} & \text{for temporal,} \\
E_{j}E_{i}^{-1}                             & \text{for spatial,}  \\
E_{j}E_{i}^{-1}T_i^{t \rightarrow t'}, t' \in \{t+1, t-1\} & \text{for spatial-temp,} \\
\end{array}
\right.
\end{array}
\end{equation}
where $E_{i}$ and $E_{j}$ are the transformation matrix from the camera coordinate system to the vehicle coordinate system. Note that this camera-to-vehicle transformation matrix is fixed for each camera across all timesteps and is relatively easy to obtain in practice, whereas the general world-to-camera extrinsics vary at every timestep and thus are costly to collect. Leveraging the fixed camera-to-vehicle transformation matrix and the camera motion predicted by the pose network, we do not require the world-to-camera extrinsics during training, which is another advantage of our method.\par
Through spatial and temporal contexts, we compute three photometric losses $\mathcal{L}_{tm}$, $\mathcal{L}_{sp}$ and $\mathcal{L}_{sp-tm}$ for each camera. We also adopt a smoothness loss~\cite{monodepth} that encourages the output depth to be locally smooth. The loss function for scale-aware localization is: 
\begin{equation}
    \mathcal{L}_{loc} = \mathcal{L}_{tm} + \lambda_{sp}\mathcal{L}_{sp} + \lambda_{sp-tm}\mathcal{L}_{sp-tm} + \lambda_{smooth}\mathcal{L}_{smooth}.
\end{equation}
By matching the small overlap between spatial and temporal contexts, the model learns scale information during training and predicts real-scale depth for single-frame input during inference, ensuring consistency across multi-frame inputs.\par
Utilizing the real scale depth map of each input image $I_i$, we obtain the position of Gaussian primitives $\mu_i$ by unprojecting the multi-view consistent depth $D_i$ into 3D space:
\begin{equation}
    \mu_i = \Pi^{-1}(I_i, D_i),
\end{equation}
where $\Pi$ is the projection matrix with camera intrinsics $K_i$ and camera-to-vehicle transformation matrix $E_i$. Hence, the scale-aware localization for Gaussian primitives is achieved.\par
\subsection{Gaussian Parameters Prediction from Single Image}\label{parameters}
In our DrivingForward, each Gaussian primitive is parameterized by properties $\{\mu, \Sigma, \alpha, c\}$ following the original 3DGS~\cite{3dgs}, where the covariance matrix $\Sigma$ can be decomposed into a scaling factor $s$ and a rotation quaternion $r$. We obtain the $\mu \in \mathbb{R}^3$ through scale-aware localization. Then we need to predict the other parameters, including scaling factor $s \in \mathbb{R}^3_+$, rotation quaternion $r \in \mathbb{R}^4$, opacity $\alpha \in [0,1]$ and color $c \in \mathbb{R}^k$ that represented by $k$ degree spherical harmonics.\par
Unlike pixelSplat~\cite{pixelsplat} and MVSplat~\cite{mvsplat} that rely on paired views to predict Gaussian parameters, we propose a Gaussian network to independently predict these parameters from each image and aggregate the Gaussian primitives across all images, which is more suitable for driving scenes with small overlap between sparse vehicle-mounted cameras. To ensure the predicted Gaussian parameters are consistent across all input views, we utilize the previous scale-aware depth and image feature from the depth network as input for the Gaussian network. Since the previous depth network learns scale information from spatial and temporal context images, we argue that the scale-aware depth and image feature can enhance the multi-view consistency of the remaining Gaussian parameters.\par
The Gaussian network $\mathcal{G}$ consists of a depth encoder $\mathcal{E}$, a feature fusion decoder $\mathcal{D}$, and four prediction heads $\mathcal{H}_{s}$, $\mathcal{H}_{r}$, $\mathcal{H}_{\alpha}$, $\mathcal{H}_{c}$ for scaling factor $s$, rotation quaternion $r$, opacity $\alpha$, and color $c$, respectively.\par
Given the output estimated depth map $D_i$ of input image $I_i$, the depth encoder $\mathcal{E}$ of the Gaussian network extracts the depth feature $F_{depth}$:
\begin{equation}
    F_{depth} = \mathcal{E}(D_i),
\end{equation}
where $F_{depth}$ contains the 3D geometric information for 2D pixels. Then the fusion decoder $\mathcal{D}$ combined $F_{depth}$ with the image feature $F_{image}$ from depth network encoder:
\begin{equation}
    F_{fusion} = \mathcal{D}(Concat(F_{depth}, F_{image})),
\end{equation}
where $Concat$ indicates the concat operation at each feature level and $F_{fusion}$ is the fusion feature for Gaussian parameters prediction. The prediction heads simply consist of two convolutions to effectively predict parameters from the fusion feature. We also apply softplus and softmax activation on $\mathcal{H}_{s}$ and $\mathcal{H}_{\alpha}$ to ensure $s \in \mathbb{R}^3_+$ and $\alpha \in [0,1]$. The parameters are obtained by:
\begin{equation}
    p = \mathcal{H}_p(F_{fusion}), \quad p \in \{s, r, \alpha, c\}.
\end{equation}
Since the Gaussian network predicts Gaussian parameters from one single frame of surround-view images without relying on additional frames, the single-frame-based prediction supports flexible single-frame or multi-frame inputs of surrounding views, rather than being restricted to fixed inputs like paired images from two adjacent frames.\par
\subsection{Joint Training Strategy}\label{training}
By applying scale-aware localization and Gaussian parameters prediction to each input view, we obtain the Gaussian primitives for all images. These primitives are then aggregated into 3D space to form a 3D representation. Novel view synthesis can be achieved through the splat-based rasterization rendering in 3DGS~\cite{3dgs}.\par
We jointly train the full model, including the depth network, the pose network, and the Gaussian network. For warp operation of the depth and pose network, we use spatial transformer network~\cite{stn} to sample the synthesis image from the source image, which is fully differentiable~\cite{monodepth}. For rendering novel views after obtaining the Gaussian primitives in 3D space, the splat-based rasterization rendering is also fully differentiable. These two operations along with other differentiable parts enable the joint training end-to-end.\par
We fuse image features from the depth network into the Gaussian network. This shared feature connects the scale-aware positions with predictions of other Gaussian parameters, allowing the Gaussian network to leverage scale information from temporal and spatial contexts. Additionally, it promotes the convergence of the full model.\par
The Gaussian network is supervised on a linear combination of L2 and LPIPS loss~\cite{lpips} between a novel view ground truth $I_{gt}$ and the rendering image $I_{render}$:
\begin{equation}
    \mathcal{L}_{render} = \beta L_2 + \gamma L_{LPIPS},
\end{equation}
with $\beta=1$ and $\gamma=0.05$. In practical training, we take the adjacent frame of input images as the novel view.
The final loss function for the full model is:
\begin{equation}
    \mathcal{L} = \mathcal{L}_{loc} + \lambda_{render}\mathcal{L}_{render}.
\end{equation}
Through the joint training strategy, we achieve scale-aware localization and Gaussian parameters prediction in one stage and support flexible multi-frame inputs, as the prediction independently depends on each frame of surrounding views.\par

\begin{table*}[t!]
\centering
\begin{tabular}{l|c|c|c|ccc}
    \toprule
    \textbf{Method} & Venue\&Year & Mode & Resolution & 
    \textbf{PSNR$\uparrow$} & \textbf{SSIM$\uparrow$} & \textbf{LPIPS$\downarrow$} \\
    \midrule
    MVSplat & ECCV 2024 & MF & 352 $\times$ 640 & 22.83 & 0.629 & 0.317 \\
    pixelSplat & CVPR 2024 & MF & 352 $\times$ 640 & \underline{25.00} & \underline{0.727} & \underline{0.298} \\
    Ours & AAAI 2025 & MF & 352 $\times$ 640 & \textbf{26.06} & \textbf{0.781} & \textbf{0.215} \\
    \midrule
    UniPad & CVPR 2024 & SF & 114 $\times$ 228 & 16.45 & 0.375 & - \\
    SelfOcc & CVPR 2024 & SF & 114 $\times$ 228 & 18.22 & 0.464 & - \\
    EmerNeRF & ICLR 2024 & SF & 114 $\times$ 228 & 20.95 & 0.585 & - \\
    DistillNeRF & NeurIPS 2024 & SF & 114 $\times$ 228 & 20.78 & 0.590 & - \\
    Ours & AAAI 2025 & SF & 352 $\times$ 640 & \underline{21.67} & \underline{0.727} & \underline{0.259} \\
    Ours & AAAI 2025 & SF & 114 $\times$ 228 & \textbf{21.76} & \textbf{0.767} & \textbf{0.194} \\
    \bottomrule
\end{tabular}
\caption{Comparison of our method against other feed-forward methods in both MF and SF mode.}
\label{tab: main}
\end{table*}
\section{Experiments}
\subsection{Experimental Setup}
\subsubsection{Implementation Details}
We implement DrivingForward with PyTorch and use a pre-built 3DGS renderer~\cite{3dgs}. We use ResNet-18 as the encoder for both the depth and pose network and a U-Net encoder for the Gaussian network. The prediction heads of the Gaussian network consist of two convolution layers. We also integrate a feature aggregate module~\cite{vfdepth} into the depth and pose network to enhance feature representation. We use an Adam optimizer with a learning rate of $1\times10^{-4}$, a batch size of 1 with 6 surround-view images as one sample. During training, we use images from the spatially adjacent left and right cameras and the temporally adjacent frames for spatial and temporal contexts. We set $\lambda_{render} = 0.01$, $\lambda_{sp} = 0.03$, $\lambda_{sp-tm}=0.1$, and $\lambda_{smooth} = 0.001$ in the loss function and train our model for 10 epochs. All experiments are conducted on a single A6000 GPU with 48GB.\par
\subsubsection{Datasets and Metrics}
The nuScenes dataset~\cite{nuscenes} is a public large-scale dataset for autonomous driving. It contains 1000 driving scenes from different geographic areas. Each scene comprises sequential frames of approximately 20 seconds, and the entire dataset encompasses around 40,000 keyframes. The images are captured by 6 vehicle-mounted cameras that cover the surrounding views. The overlap between adjacent camera images is minimal, at approximately 10\%. In our experiments, we use the default split of 700 scenes for training and 150 scenes for validation. Unless otherwise specified, we adopt a default resolution of $352 \times 640$. To evaluate the quality of reconstructed scenes, we synthesize novel views for frames adjacent to the input frame images and compute the peak signal-to-noise ratio (PSNR), structural similarity index (SSIM)~\cite{ssim}, and perceptual distance (LPIPS)~\cite{lpips}.\par
\subsubsection{Baselines}
Since our method is among the first to explore the real-time reconstruction of driving scenes, no existing benchmarks are available. Therefore, we defined two modes of novel view synthesis to accommodate different comparison methods. The first is the \textbf{single-frame (SF) mode}, where given one single frame of surrounding views at timestep $t$, the goal is to synthesize the next frame's surround-view images at timestep $t+1$. The other mode is the \textbf{multi-frame (MF) mode}, where given two surround-view images of interval frames, i.e. the surround-view images at timestep $t-1$ and $t+1$, the goal is to synthesize the intermediate surround-view images at the timestep $t$. Using the two modes of novel view synthesis, we compare our method against both feed-forward and scene-optimized reconstruction methods.\par
For feed-forward methods, we compare our model with MVSplat~\cite{mvsplat}, pixelSplat~\cite{pixelsplat}, and DistllNeRF~\cite{distillnerf} (along with its comparison methods EmerNeRF~\cite{emernerf}, Unipad~\cite{unipad}, and SelfOcc~\cite{selfocc}). MVSplat and pixelSplat are designed for training on datasets with densely overlapping inputs. Given that temporally adjacent frames have significantly more overlap than spatially adjacent frames, we use MF mode to meet their overlapping requirements. We retrained them by combining their publicly available codebases with our dataset and data loader. DistillNeRF does not release the code and cannot be trained. To enable a fair comparison, we align our method with the settings specified in its paper, including input, rendered novel view, image resolution, and validation scenes. Since SF mode aligns with DistillNeRF's input and novel views, we train our model using this mode.\par
For scene-optimized reconstruction methods, we compare our trained method to the original 3DGS~\cite{3dgs} using the SF mode. Note that our method does not require test-time optimization while 3DGS needs to be optimized scene by scene. We conduct this comparison to demonstrate that our method can achieve comparable or even higher reconstruction quality without test-time optimization.\par 
\begin{figure*}[t!]
    \centering
    \includegraphics[width=.9\textwidth]{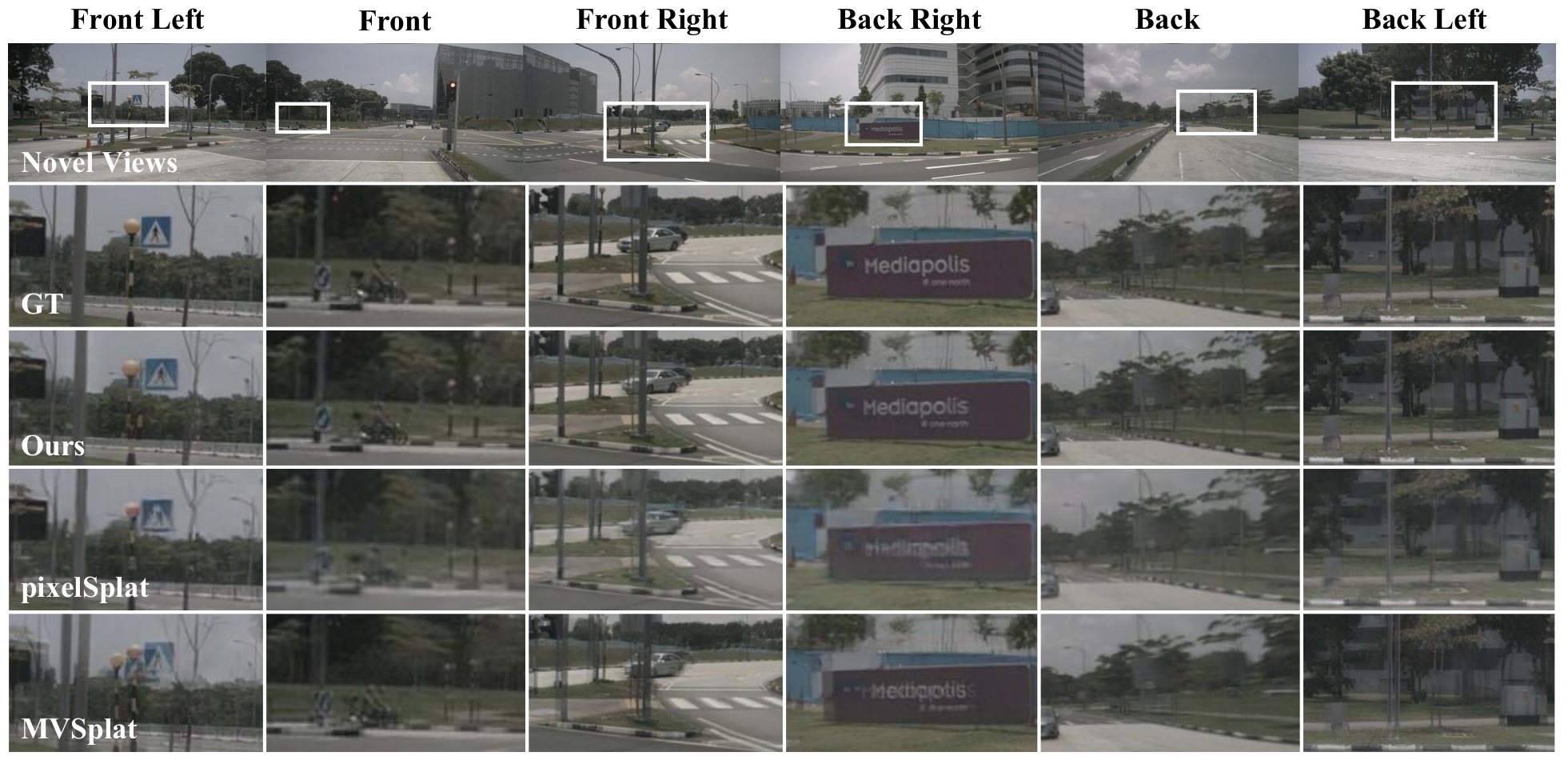}
    \caption{Qualitative results of novel surrounding views. Details from surrounding views are present for easy comparison. }
    \label{fig: vis}
\end{figure*}
\subsection{Comparison with Feed-forward Methods}
We report the quantitative results of comparison with state-of-the-art feed-forward methods in Table~\ref{tab: main}. 
Although we adapt our method to align with the different settings of the baselines, we outperform them across all metrics under the corresponding configuration. The qualitative comparisons of MVSplat, pixelSplat, and our method are visualized in Figure~\ref{fig: vis}. Our DrivingForward achieves the highest quality on novel view results even for challenging details, such as the traffic sign in the front left view and the monument with words in the back right view. Other methods show obvious artifacts in these regions, while our method synthesizes clear novel views without such artifacts.\par
\begin{table}[t!]
    \centering
    \begin{tabular}{l|c|ccc}
    \toprule
    \multirow{2}{*}{Method} & \multirow{2}{*}{\shortstack{Test Time \\ (per scene)}} & \multirow{2}{*}{PSNR$\uparrow$} & \multirow{2}{*}{SSIM$\uparrow$} & \multirow{2}{*}{LPIPS$\downarrow$} \\
    & & & & \\
    \midrule
    3DGS & $\approx$ 9 min & 19.57 & 0.599 & 0.465 \\
    Ours & 0.29 s & 21.84 & 0.758 & 0.181 \\
    \bottomrule
    \end{tabular}
    \caption{Comparison of our feed-forward method against scene-optimized 3DGS in SF mode.}
    \label{tab: 3dgs}
\end{table}
\begin{table}[t!]
    \centering
    \small
    \begin{tabular}{l|cc|cc}
    \toprule
    \multirow{3}{*}{Method} & \multicolumn{2}{c}{Time} & \multicolumn{2}{c}{Memory} \\
     & \multirow{2}{*}{\shortstack{Training \\ (total)}} & \multirow{2}{*}{\shortstack{Inference \\ (per scene)}} & \multirow{2}{*}{Training} & \multirow{2}{*}{Inference} \\
    & & & & \\
    \midrule
    MVSplat & $\approx$ 4 days & 1.39 s & 35.3 GB & 5.45 GB \\
    pixelSplat & $\approx$ 6 days & 2.95 s & 45.4 GB & 11.3 GB \\
    Ours & $\approx$ 3 days & 0.63 s & 40.0 GB & 7.58 GB\\
    \bottomrule
    \end{tabular}
    \caption{Comparison of our method against MVSplat and pixelSplat on runtime and GPU memory usage in MF mode. }
    \label{tab: time}
\end{table}
\subsection{Comparison with Scene-optimized Methods}
We compare our feed-forward method against the original 3DGS which represents the scene-optimized methods. In SF mode, we train our model and select the first three scenes from the validation set. Then we optimize the 3DGS model for each scene individually and the rendered novel view images of 3DGS models are compared with ours. The average test time pre-scene and metrics across the three scenes are reported in Table~\ref{tab: 3dgs}. 3DGS takes several minutes to synthesize the novel views of a scene (6 images). In contrast, our feed-forward method completes this within half a second and achieves higher reconstruction quality without the costly test-time optimization.\par
\subsection{Comparison on Runtime and Memory}
We compare the runtime and GPU memory usage of our method against MVSplat and pixelSplat under the MF mode. As shown in Table~\ref{tab: time}, our method requires less training time and achieves faster inference speed in synthesizing novel surrounding views that are composed of 6 images with a resolution of 352 $\times$ 640. We also report the GPU memory usage in practical training and inference. Note that our GPU memory usage is not the lowest as we use two frames of surrounding images in one batch, while MVSplat and pixelSplat use two images in one batch. The difference is due to different training architectures. Despite this, our memory usage remains comparable to that of other methods. \par
\begin{table}[t!]
\centering
\begin{tabular}{l|ccc}
    \toprule
    Model & PSNR$\uparrow$ & SSIM$\uparrow$ & LPIPS$\downarrow$ \\
    \midrule
    Full Model & 21.67 & 0.727 & 0.259 \\
    w/o Joint Training & 21.50 & 0.726 & 0.261 \\
    w/o Image Feature Share & 19.48 & 0.699 & 0.314\\
    w/o Scale-aware Loc. & 11.96 & 0.466 & 0.772 \\
    \bottomrule
\end{tabular}
\caption{Ablation studies. All ablation models are based on our full model by removing the corresponding module.}
\label{tab: ablation}
\end{table}
\subsection{Ablation Studies}
We conduct ablation experiments to demonstrate the effectiveness of our method in detail. We train our model using the single-frame (SF) mode of novel view synthesis and refer to it as ``Full Model'' in Table~\ref{tab: ablation}.\par
To validate the effectiveness of the joint training strategy, we first train the scale-aware localization module, which includes only the depth and pose networks. We then use this pre-trained model to train the Gaussian network (2nd row, w/o Joint Training). The result in Table~\ref{tab: ablation} shows that joint training enables end-to-end training without decreasing performance and even gains a better performance. This indicates that the joint training is able to facilitate model convergence and can lead to improvement.\par
In addition to the estimated depth map, the shared image feature from the depth network is another key component for connecting to the subsequent Gaussian network. To demonstrate the importance of this connection, we ablate the shared image feature and instead construct another feature extractor like the depth network encoder for the Gaussian network. As illustrated in Table~\ref{tab: ablation}, the model ``w/o Image Feature Share'' (3rd row) drops more than 2 dB PSNR. This highlights the importance of the shared image feature in connecting the positions and other parameters of Gaussian primitives.\par
To investigate whether the scale-aware localization is necessary, we replace the scale-aware localization with a monocular depth estimation model~\cite{monodepth2} (4th row, w/o Scale-aware Loc.), as shown in Table~\ref{tab: ablation}. Although monocular depth estimation models can estimate depth from a single image, they are scale-ambiguous, which indicates the depth maps are inconsistent across different views and lead to a large drop in performance. This demonstrates the irreplaceable role of scale-aware localization.\par
\section{Conclusion}
We introduce DrivingForward, a feed-forward Gaussian Splatting model that achieves real-time driving scene reconstruction from flexible surround-view input. To solve the problem of minimal overlap of the surrounding views, we predict the Gaussian primitives from each image independently and propose the scale-aware localization to obtain the multi-view consistent position for Gaussian primitives. A Gaussian network predicts primitives' other parameters from each image. The individual prediction enables flexible multi-frame inputs of surrounding views. Our method does not require depth ground truth and is extrinsic-free during training. At the inference stage, our model is faster than other methods and achieves higher reconstruction quality for driving scenes compared with both existing feed-forward and scene-optimized reconstruction methods. In the future, this work is expected to integrate with human perception research~\cite{futurehuman} to develop a more intelligent human-in-scene perception system. \par

\section*{Acknowledgements}
This work was supported in part by the National Natural Science Foundation of China (62302167, U23A20343, 62222602, 62176092, 62472282, 72192821); in part by Shanghai Sailing Program (23YF1410500); in part by Chenguang Program of Shanghai Education Development Foundation and Shanghai Municipal Education Commission (23CGA34); in part by the Fundamental Research Funds for the Central Universities (YG2023QNA35); in part by YuCaiKe [2023] (231111310300).

\bibliography{xxxx}

\begin{thebibliography}{41}
\providecommand{\natexlab}[1]{#1}

\bibitem[{Caesar et~al.(2020)Caesar, Bankiti, Lang, Vora, Liong, Xu, Krishnan, Pan, Baldan, and Beijbom}]{nuscenes}
Caesar, H.; Bankiti, V.; Lang, A.~H.; Vora, S.; Liong, V.~E.; Xu, Q.; Krishnan, A.; Pan, Y.; Baldan, G.; and Beijbom, O. 2020.
\newblock nuscenes: A multimodal dataset for autonomous driving.
\newblock In \emph{Proceedings of the IEEE/CVF Conference on Computer Vision and Pattern Recognition (CVPR)}, 11621--11631.

\bibitem[{Cai et~al.(2023)Cai, Pan, Yao, Ngo, and Mei}]{3d_detection_objectfusion}
Cai, Q.; Pan, Y.; Yao, T.; Ngo, C.-W.; and Mei, T. 2023.
\newblock Objectfusion: Multi-modal 3d object detection with object-centric fusion.
\newblock In \emph{Proceedings of the IEEE/CVF International Conference on Computer Vision (ICCV)}, 18067--18076.

\bibitem[{Charatan et~al.(2024)Charatan, Li, Tagliasacchi, and Sitzmann}]{pixelsplat}
Charatan, D.; Li, S.~L.; Tagliasacchi, A.; and Sitzmann, V. 2024.
\newblock pixelsplat: 3d gaussian splats from image pairs for scalable generalizable 3d reconstruction.
\newblock In \emph{Proceedings of the IEEE/CVF Conference on Computer Vision and Pattern Recognition (CVPR)}, 19457--19467.

\bibitem[{Chen et~al.(2023)Chen, Zhang, Wang, Wang, and Zhao}]{3d_detection_futr3d}
Chen, X.; Zhang, T.; Wang, Y.; Wang, Y.; and Zhao, H. 2023.
\newblock Futr3d: A unified sensor fusion framework for 3d detection.
\newblock In \emph{proceedings of the IEEE/CVF Conference on Computer Vision and Pattern Recognition (CVPR)}, 172--181.

\bibitem[{Chen et~al.(2024)Chen, Xu, Zheng, Zhuang, Pollefeys, Geiger, Cham, and Cai}]{mvsplat}
Chen, Y.; Xu, H.; Zheng, C.; Zhuang, B.; Pollefeys, M.; Geiger, A.; Cham, T.-J.; and Cai, J. 2024.
\newblock Mvsplat: Efficient 3d gaussian splatting from sparse multi-view images.
\newblock \emph{arXiv preprint arXiv:2403.14627}.

\bibitem[{Du et~al.(2023)Du, Smith, Tewari, and Sitzmann}]{du}
Du, Y.; Smith, C.; Tewari, A.; and Sitzmann, V. 2023.
\newblock Learning to render novel views from wide-baseline stereo pairs.
\newblock In \emph{Proceedings of the IEEE/CVF Conference on Computer Vision and Pattern Recognition (CVPR)}, 4970--4980.

\bibitem[{Godard, Mac~Aodha, and Brostow(2017)}]{monodepth}
Godard, C.; Mac~Aodha, O.; and Brostow, G.~J. 2017.
\newblock Unsupervised monocular depth estimation with left-right consistency.
\newblock In \emph{Proceedings of the IEEE/CVF Conference on Computer Vision and Pattern Recognition (CVPR)}, 270--279.

\bibitem[{Godard et~al.(2019)Godard, Mac~Aodha, Firman, and Brostow}]{monodepth2}
Godard, C.; Mac~Aodha, O.; Firman, M.; and Brostow, G.~J. 2019.
\newblock Digging into self-supervised monocular depth estimation.
\newblock In \emph{Proceedings of the IEEE/CVF International Conference on Computer Vision (ICCV)}, 3828--3838.

\bibitem[{Guizilini et~al.(2022)Guizilini, Vasiljevic, Ambrus, Shakhnarovich, and Gaidon}]{fsm}
Guizilini, V.; Vasiljevic, I.; Ambrus, R.; Shakhnarovich, G.; and Gaidon, A. 2022.
\newblock Full surround monodepth from multiple cameras.
\newblock \emph{IEEE Robotics and Automation Letters (RAL)}, 7(2): 5397--5404.

\bibitem[{Guo et~al.(2024)Guo, Li, Hu, Zhang, and Wang}]{futurehuman}
Guo, D.; Li, K.; Hu, B.; Zhang, Y.; and Wang, M. 2024.
\newblock Benchmarking micro-action recognition: dataset, method, and application.
\newblock \emph{IEEE Transactions on Circuits and Systems for Video Technology (TCSVT)}, 34(7): 6238--6252.

\bibitem[{Guo et~al.(2023)Guo, Deng, Li, Bai, Shi, Wang, Ding, Wang, and Li}]{streetsurf}
Guo, J.; Deng, N.; Li, X.; Bai, Y.; Shi, B.; Wang, C.; Ding, C.; Wang, D.; and Li, Y. 2023.
\newblock Streetsurf: Extending multi-view implicit surface reconstruction to street views.
\newblock \emph{arXiv preprint arXiv:2306.04988}.

\bibitem[{Huang et~al.(2024{\natexlab{a}})Huang, Wei, Zheng, An, Lu, Zhan, Tomizuka, Keutzer, and Zhang}]{s3gaussian}
Huang, N.; Wei, X.; Zheng, W.; An, P.; Lu, M.; Zhan, W.; Tomizuka, M.; Keutzer, K.; and Zhang, S. 2024{\natexlab{a}}.
\newblock S3Gaussian: Self-Supervised Street Gaussians for Autonomous Driving.
\newblock \emph{arXiv preprint arXiv:2405.20323}.

\bibitem[{Huang et~al.(2024{\natexlab{b}})Huang, Zheng, Zhang, Zhou, and Lu}]{selfocc}
Huang, Y.; Zheng, W.; Zhang, B.; Zhou, J.; and Lu, J. 2024{\natexlab{b}}.
\newblock Selfocc: Self-supervised vision-based 3d occupancy prediction.
\newblock In \emph{Proceedings of the IEEE/CVF Conference on Computer Vision and Pattern Recognition (CVPR)}, 19946--19956.

\bibitem[{Jaderberg et~al.(2015)Jaderberg, Simonyan, Zisserman et~al.}]{stn}
Jaderberg, M.; Simonyan, K.; Zisserman, A.; et~al. 2015.
\newblock Spatial transformer networks.
\newblock \emph{Advances in Neural Information Processing Systems (NeurIPS)}, 28.

\bibitem[{Ji et~al.(2024)Ji, Zhu, Tang, Liu, Zhang, Xie, and Tan}]{scene_fastlgs}
Ji, Y.; Zhu, H.; Tang, J.; Liu, W.; Zhang, Z.; Xie, Y.; and Tan, X. 2024.
\newblock FastLGS: Speeding up Language Embedded Gaussians with Feature Grid Mapping.
\newblock In \emph{Proceedings of the AAAI Conference on Artificial Intelligence.}

\bibitem[{Kerbl et~al.(2023)Kerbl, Kopanas, Leimk{\"{u}}hler, and Drettakis}]{3dgs}
Kerbl, B.; Kopanas, G.; Leimk{\"{u}}hler, T.; and Drettakis, G. 2023.
\newblock 3D Gaussian Splatting for Real-Time Radiance Field Rendering.
\newblock \emph{{ACM} Trans. Graph.}, 42(4): 139:1--139:14.

\bibitem[{Khan et~al.(2024)Khan, Fazlali, Sharma, Cao, Bai, Ren, and Liu}]{autosplat}
Khan, M.; Fazlali, H.; Sharma, D.; Cao, T.; Bai, D.; Ren, Y.; and Liu, B. 2024.
\newblock AutoSplat: Constrained Gaussian Splatting for Autonomous Driving Scene Reconstruction.
\newblock \emph{arXiv preprint arXiv:2407.02598}.

\bibitem[{Kim et~al.(2022)Kim, Hur, Nguyen, and Jeong}]{vfdepth}
Kim, J.-H.; Hur, J.; Nguyen, T.~P.; and Jeong, S.-G. 2022.
\newblock Self-supervised surround-view depth estimation with volumetric feature fusion.
\newblock \emph{Advances in Neural Information Processing Systems (NeurIPS)}, 35: 4032--4045.

\bibitem[{Li et~al.(2024)Li, Zhang, Bai, Zheng, Ning, Zhou, and Gu}]{dngaussian}
Li, J.; Zhang, J.; Bai, X.; Zheng, J.; Ning, X.; Zhou, J.; and Gu, L. 2024.
\newblock Dngaussian: Optimizing sparse-view 3d gaussian radiance fields with global-local depth normalization.
\newblock In \emph{Proceedings of the IEEE/CVF Conference on Computer Vision and Pattern Recognition (CVPR)}, 20775--20785.

\bibitem[{Li et~al.(2022{\natexlab{a}})Li, Wang, Wang, and Zhao}]{online_mapping_hdmapnet}
Li, Q.; Wang, Y.; Wang, Y.; and Zhao, H. 2022{\natexlab{a}}.
\newblock HDMapNet: An Online HD Map Construction and Evaluation Framework.
\newblock In \emph{International Conference on Robotics and Automation (ICRA)}, 4628--4634.

\bibitem[{Li et~al.(2022{\natexlab{b}})Li, Wang, Li, Xie, Sima, Lu, Qiao, and Dai}]{bev_bevformer}
Li, Z.; Wang, W.; Li, H.; Xie, E.; Sima, C.; Lu, T.; Qiao, Y.; and Dai, J. 2022{\natexlab{b}}.
\newblock Bevformer: Learning bird’s-eye-view representation from multi-camera images via spatiotemporal transformers.
\newblock In \emph{European Conference on Computer Vision (ECCV)}, 1--18.

\bibitem[{Liang et~al.(2022)Liang, Xie, Yu, Xia, Lin, Wang, Tang, Wang, and Tang}]{bev_bevfusion1}
Liang, T.; Xie, H.; Yu, K.; Xia, Z.; Lin, Z.; Wang, Y.; Tang, T.; Wang, B.; and Tang, Z. 2022.
\newblock Bevfusion: A simple and robust lidar-camera fusion framework.
\newblock \emph{Advances in Neural Information Processing Systems (NeurIPS)}, 35: 10421--10434.

\bibitem[{Liu et~al.(2023)Liu, Tang, Amini, Yang, Mao, Rus, and Han}]{bev_bevfusion2}
Liu, Z.; Tang, H.; Amini, A.; Yang, X.; Mao, H.; Rus, D.~L.; and Han, S. 2023.
\newblock Bevfusion: Multi-task multi-sensor fusion with unified bird's-eye view representation.
\newblock In \emph{IEEE International Conference on Robotics and Automation (ICRA)}, 2774--2781.

\bibitem[{Ma et~al.(2024)Ma, Tan, Qu, Ma, Zhang, and Xie}]{scene_cotr}
Ma, Q.; Tan, X.; Qu, Y.; Ma, L.; Zhang, Z.; and Xie, Y. 2024.
\newblock Cotr: Compact occupancy transformer for vision-based 3d occupancy prediction.
\newblock In \emph{Proceedings of the IEEE/CVF Conference on Computer Vision and Pattern Recognition (CVPR)}, 19936--19945.

\bibitem[{Mildenhall et~al.(2021)Mildenhall, Srinivasan, Tancik, Barron, Ramamoorthi, and Ng}]{nerf}
Mildenhall, B.; Srinivasan, P.~P.; Tancik, M.; Barron, J.~T.; Ramamoorthi, R.; and Ng, R. 2021.
\newblock Nerf: Representing scenes as neural radiance fields for view synthesis.
\newblock \emph{Communications of the ACM}, 65(1): 99--106.

\bibitem[{Suhail et~al.(2022)Suhail, Esteves, Sigal, and Makadia}]{gpnr}
Suhail, M.; Esteves, C.; Sigal, L.; and Makadia, A. 2022.
\newblock Generalizable patch-based neural rendering.
\newblock In \emph{European Conference on Computer Vision (ECCV)}, 156--174.

\bibitem[{Sun et~al.(2024{\natexlab{a}})Sun, Zhang, Tan, Peng, Qu, and Xie}]{scene_unitomulti}
Sun, T.; Zhang, Z.; Tan, X.; Peng, Y.; Qu, Y.; and Xie, Y. 2024{\natexlab{a}}.
\newblock Uni-to-Multi Modal Knowledge Distillation for Bidirectional LiDAR-Camera Semantic Segmentation.
\newblock \emph{IEEE Transactions on Pattern Analysis and Machine Intelligence (TPAMI)}, 46(12): 11059--11072.

\bibitem[{Sun et~al.(2024{\natexlab{b}})Sun, Zhang, Tan, Qu, and Xie}]{scene_imageunderstands}
Sun, T.; Zhang, Z.; Tan, X.; Qu, Y.; and Xie, Y. 2024{\natexlab{b}}.
\newblock Image understands point cloud: Weakly supervised 3D semantic segmentation via association learning.
\newblock \emph{IEEE Transactions on Image Processing (TIP)}, 33: 1838--1852.

\bibitem[{Szymanowicz, Rupprecht, and Vedaldi(2024)}]{splatter}
Szymanowicz, S.; Rupprecht, C.; and Vedaldi, A. 2024.
\newblock Splatter image: Ultra-fast single-view 3d reconstruction.
\newblock In \emph{Proceedings of the IEEE/CVF Conference on Computer Vision and Pattern Recognition (CVPR)}, 10208--10217.

\bibitem[{Tan et~al.(2023)Tan, Ma, Gong, Xu, Zhang, Song, Qu, Xie, and Ma}]{scene_positive-negative}
Tan, X.; Ma, Q.; Gong, J.; Xu, J.; Zhang, Z.; Song, H.; Qu, Y.; Xie, Y.; and Ma, L. 2023.
\newblock Positive-negative receptive field reasoning for omni-supervised 3d segmentation.
\newblock \emph{IEEE Transactions on Pattern Analysis and Machine Intelligence (TPAMI)}, 45(12): 15328--15344.

\bibitem[{Wang et~al.(2024)Wang, Kim, Yang, Yu, Ivanovic, Waslander, Wang, Fidler, Pavone, and Karkus}]{distillnerf}
Wang, L.; Kim, S.~W.; Yang, J.; Yu, C.; Ivanovic, B.; Waslander, S.~L.; Wang, Y.; Fidler, S.; Pavone, M.; and Karkus, P. 2024.
\newblock DistillNeRF: Perceiving 3D Scenes from Single-Glance Images by Distilling Neural Fields and Foundation Model Features.
\newblock \emph{Advances in Neural Information Processing Systems (NeurIPS)}.

\bibitem[{Wang et~al.(2004)Wang, Bovik, Sheikh, and Simoncelli}]{ssim}
Wang, Z.; Bovik, A.~C.; Sheikh, H.~R.; and Simoncelli, E.~P. 2004.
\newblock Image quality assessment: from error visibility to structural similarity.
\newblock \emph{IEEE Transactions on Image Processing (TIP)}, 13(4): 600--612.

\bibitem[{Wei et~al.(2023)Wei, Zhao, Zheng, Zhu, Rao, Huang, Lu, and Zhou}]{surrounddepth}
Wei, Y.; Zhao, L.; Zheng, W.; Zhu, Z.; Rao, Y.; Huang, G.; Lu, J.; and Zhou, J. 2023.
\newblock Surrounddepth: Entangling surrounding views for self-supervised multi-camera depth estimation.
\newblock In \emph{Conference on Robot Learning (CoRL)}, 539--549.

\bibitem[{Xiong et~al.(2023)Xiong, Muttukuru, Upadhyay, Chari, and Kadambi}]{sparsegs}
Xiong, H.; Muttukuru, S.; Upadhyay, R.; Chari, P.; and Kadambi, A. 2023.
\newblock Sparsegs: Real-time 360 {\textdegree} sparse view synthesis using gaussian splatting.
\newblock \emph{arXiv preprint arXiv:2312.00206}.

\bibitem[{Yan et~al.(2024)Yan, Lin, Zhou, Wang, Sun, Zhan, Lang, Zhou, and Peng}]{streetgaussian}
Yan, Y.; Lin, H.; Zhou, C.; Wang, W.; Sun, H.; Zhan, K.; Lang, X.; Zhou, X.; and Peng, S. 2024.
\newblock Street Gaussians: Modeling Dynamic Urban Scenes with Gaussian Splatting.
\newblock In \emph{European Conference on Computer Vision (ECCV)}, volume 15131, 156--173.

\bibitem[{Yang et~al.(2024{\natexlab{a}})Yang, Zhang, Huang, Wu, Zhu, He, Tang, Zhao, Qiu, Lin et~al.}]{unipad}
Yang, H.; Zhang, S.; Huang, D.; Wu, X.; Zhu, H.; He, T.; Tang, S.; Zhao, H.; Qiu, Q.; Lin, B.; et~al. 2024{\natexlab{a}}.
\newblock Unipad: A universal pre-training paradigm for autonomous driving.
\newblock In \emph{Proceedings of the IEEE/CVF Conference on Computer Vision and Pattern Recognition (CVPR)}, 15238--15250.

\bibitem[{Yang et~al.(2024{\natexlab{b}})Yang, Ivanovic, Litany, Weng, Kim, Li, Che, Xu, Fidler, Pavone, and Wang}]{emernerf}
Yang, J.; Ivanovic, B.; Litany, O.; Weng, X.; Kim, S.~W.; Li, B.; Che, T.; Xu, D.; Fidler, S.; Pavone, M.; and Wang, Y. 2024{\natexlab{b}}.
\newblock EmerNeRF: Emergent spatial-temporal scene decomposition via self-supervision.
\newblock In \emph{International Conference on Learning Representations (ICLR)}.

\bibitem[{Yu et~al.(2021)Yu, Ye, Tancik, and Kanazawa}]{pixelnerf}
Yu, A.; Ye, V.; Tancik, M.; and Kanazawa, A. 2021.
\newblock pixelnerf: Neural radiance fields from one or few images.
\newblock In \emph{Proceedings of the IEEE/CVF Conference on Computer Vision and Pattern Recognition (CVPR)}, 4578--4587.

\bibitem[{Zhang et~al.(2018)Zhang, Isola, Efros, Shechtman, and Wang}]{lpips}
Zhang, R.; Isola, P.; Efros, A.~A.; Shechtman, E.; and Wang, O. 2018.
\newblock The unreasonable effectiveness of deep features as a perceptual metric.
\newblock In \emph{Proceedings of the IEEE/CVF Conference on Computer Vision and Pattern Recognition (CVPR)}, 586--595.

\bibitem[{Zheng et~al.(2024)Zheng, Zhou, Shao, Liu, Zhang, Nie, and Liu}]{gps-gaussian}
Zheng, S.; Zhou, B.; Shao, R.; Liu, B.; Zhang, S.; Nie, L.; and Liu, Y. 2024.
\newblock Gps-gaussian: Generalizable pixel-wise 3d gaussian splatting for real-time human novel view synthesis.
\newblock In \emph{Proceedings of the IEEE/CVF Conference on Computer Vision and Pattern Recognition (CVPR)}, 19680--19690.

\bibitem[{Zhou et~al.(2024)Zhou, Lin, Shan, Wang, Sun, and Yang}]{drivinggaussian}
Zhou, X.; Lin, Z.; Shan, X.; Wang, Y.; Sun, D.; and Yang, M.-H. 2024.
\newblock Drivinggaussian: Composite gaussian splatting for surrounding dynamic autonomous driving scenes.
\newblock In \emph{Proceedings of the IEEE/CVF Conference on Computer Vision and Pattern Recognition (CVPR)}, 21634--21643.

\end{thebibliography}

\newpage
\clearpage

\begin{center}
\LARGE\bf
    Supplementary Material\par
\vspace{20pt}
\end{center}
\renewcommand\thesection{\Alph{section}}
\setcounter{section}{0}
\setcounter{figure}{0}

\section{Training Details}
We use the same nuScenes dataset~\cite{nuscenes} and the default scene split (700 scenes for training and 150 scenes for validation) across all experiments in both the main paper and supplementary material. For frames in the training scenes, all models are trained on the same frames following Wei~\shortcite{surrounddepth}. For frames in the validation scenes, the MF mode uses the validation frames from Wei~\shortcite{surrounddepth}, excluding ones without the previous or next frame. The SF mode uses the validation frames from DistillNeRF~\cite{distillnerf} to align with DistillNeRF and its comparison methods. Except for comparisons with DistillNeRF and its related methods, where the resolution is set to 114 $\times$ 228, all the other experiments are conducted at a resolution of 352 $\times$ 640.\par
\subsection{pixelSplat}
Following the original settings of pixelSplat~\cite{pixelsplat}, we train it for 300,000 iterations with a batch size of 1 and the authors' default hyperparameters. To adapt it for our driving scene dataset, we apply minimal modifications to the data loader: In MF mode, we use two interval frames from a single camera as input to predict intermediate frame images and synthesize surrounding views by predicting from all cameras sequentially; in SF mode, we use images from two adjacent cameras to predict the next frame and also synthesize surrounding views by predicting from cameras sequentially.\par
\par
\subsection{MVSplat}
Following the original settings of MVSplat~\cite{mvsplat}, we train it for 300,000 iterations with a batch size of 2 and the authors' default hyperparameters. Similar to pixelSplat, we also adjust the data loader to adapt to driving scenes with the MF mode and SF mode.

\section{More Experiment Results and Analysis}
In this section, we provide more experimental results and analysis to supplement the experiments in the main paper.\par
\subsection{Comparison with pixelSplat and MVSplat in SF mode} 
In the main paper, we compare our DrivingForward with pixelSplat and MVSplat in MF mode. The reason we use MF mode instead of SF mode has been explained in the main paper: both MVSplat and pixelSplat are designed to be trained with densely overlapping input views and temporally adjacent frames in MF mode offer significantly more overlap than spatially adjacent frames in SF mode.\par
To demonstrate that pixelSplat and MVSplat indeed fail to work well in SF mode, we compare our DrivingForward with pixelSplat and MVSplat using SF mode in Table~\ref{tab: main_SF}. The results show that our method significantly outperforms the pixelSplat and MVSplat and achieves reasonable across all metrics. While pixelSplat and MVSplat exhibit extremely poor performance on each metric, indicating that these methods are not effective in the SF mode. 
Therefore, we use the MF mode in the main paper, which is better suited for pixelSplat and MVSplat. Despite this, our method outperforms them in both the SF and MF modes.\par
\begin{figure*}[t!]
    \centering    \includegraphics[width=0.95\textwidth]{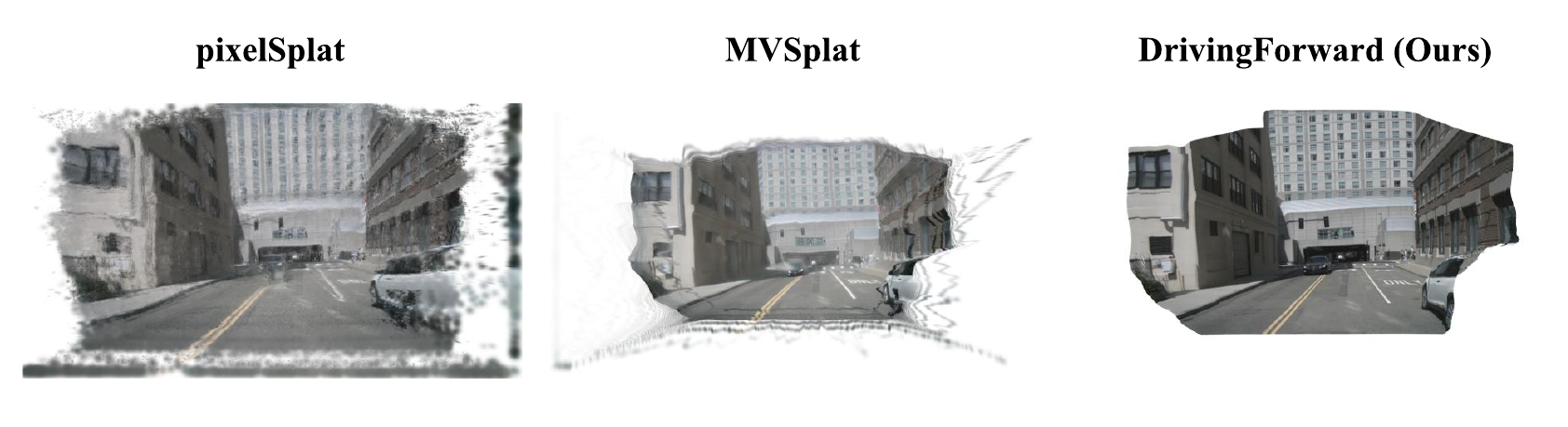}
    \caption{Visualization Comparison of Gaussian Primitives. We compare the reconstructed geometry quality by visualizing zoom-out views of 3D Gaussians Primitives predicted by pixelSplat, MVSplat, and our DrivingForward. Unlike pixelSplat and MVSplat exhibit obvious floating artifacts at the boundaries and are blurred inside the scene, our DrivingForward maintains clear edges and high quality of inside Gaussian primitives, demonstrating its effectiveness in driving scenes.}
    \label{fig: gaussian}
\end{figure*}
\begin{table*}[t!]
\centering
\begin{tabular}{l|c|c|c|ccc}
    \toprule
    \textbf{Method} & Venue\&Year & Mode & Resolution & 
    \textbf{PSNR$\uparrow$} & \textbf{SSIM$\uparrow$} & \textbf{LPIPS$\downarrow$} \\
    \midrule
    pixelSplat & CVPR 2024 & SF & 352 $\times$ 640 & 17.54 & 0.502 & 0.519 \\
    MVSplat & ECCV 2024 & SF & 352 $\times$ 640 & 17.57 & 0.528 & 0.477 \\
    Ours & AAAI 2025 & SF & 352 $\times$ 640 & 21.67 & 0.727 & 0.259 \\
    \bottomrule
\end{tabular}
\caption{Comparison of our DrivingForward with pixelSplat and MVSplat in SF mode. In the main paper, we compare our DrivingForward with pixelSplat and MVSplat in MF mode to adapt their requirement of densely overlapping inputs, and our method outperforms them. In SF mode, our performance exceeds pixelSplat and MVSplat by a larger margin, proving the fairness of the comparison in the main paper and further demonstrating the effectiveness of our method.}
\label{tab: main_SF}
\end{table*}
\subsection{Comparison with 3DGS in MF mode}
In the main paper, we compare our DrivingForward with 3DGS in an SF mode. We select the first three scenes in the validation set, taking the first frame of each scene as input, and the second frame images of each scene as novel views. In this SF mode, our method outperforms 3DGS.\par
To further demonstrate the advancement of our method, we also compare our DrvingForward with 3DGS in MF mode, which provides a more dense input that is required by 3DGS. We select the same first three scenes in the validation set, taking the first and third frames as input and the second frame images as novel views. As shown in Table~\ref{tab: 3dgs_MF}, Our method also achieves better results than 3DGS in MF mode, while only taking less than one second to infer, without any test-time optimization.\par
\begin{table}[t!]
    \centering
    \begin{tabular}{l|c|ccc}
    \toprule
    \multirow{2}{*}{Method} & \multirow{2}{*}{\shortstack{Test Time \\ (per scene)}} & \multirow{2}{*}{PSNR$\uparrow$} & \multirow{2}{*}{SSIM$\uparrow$} & \multirow{2}{*}{LPIPS$\downarrow$} \\
    & & & & \\
    \midrule
    3DGS & $\approx$ 9 min & 25.87 & 0.682 & 0.342 \\
    Ours & 0.63 s & 26.66 & 0.803 & 0.188 \\
    \bottomrule
    \end{tabular}
    \caption{Comparison of our feed-forward method against scene-optimized 3DGS in MF mode. Combining the main paper results in SF mode, our DrivingForward consistently outperforms across all metrics and synthesizes a scene of 6 images within one second. This comparison highlights our method's ability to achieve real-time inference without test-time optimization while maintaining high reconstruction quality.}
    \label{tab: 3dgs_MF}
\end{table}
\subsection{Flexibility of Our DrivingForward}
The difference between SF mode and MF lies in the number of input frames. In SF mode, surround-view images of one frame are input to predict the next frame of surround-view images, while MF mode inputs two interval surround-view frames to synthesize the intermediate surround-view frame. The different number of input frames leads to different overlapping of input views. For SF mode, since only one frame of each surrounding view is input, the overlapping only exists in the views of spatially adjacent cameras, which is minimal. For MF mode, since two interval frames of surrounding views are input, the two interval frames of the same camera have a relatively large overlap.\par
For MVSplat and pixelSplat that require densely overlapping input images, they are only suitable for MF mode with large overlaps and perform extremely poorly under SF mode. For DistillNeRF, its paper only shows the performance in SF mode. In contrast, our DrivingForward not only adapts to both SF and MF modes but also outperforms other methods in their corresponding mode, indicating its flexibility and effectiveness.\par

\section{More Visualization Results}
In this section, we provide more visualization results for further comparison with other methods.\par

\subsection{Visualization of Gaussian Primitives}
We visualize the Gaussian Primitives predicted by MVSplat, pixelSplat, and our method in MF mode, as shown in Figure~\ref{fig: gaussian}.\par 

\subsection{Visualization of Novel Views}
\begin{figure*}[!t]
    \centering
    \includegraphics[width=0.95\textwidth]{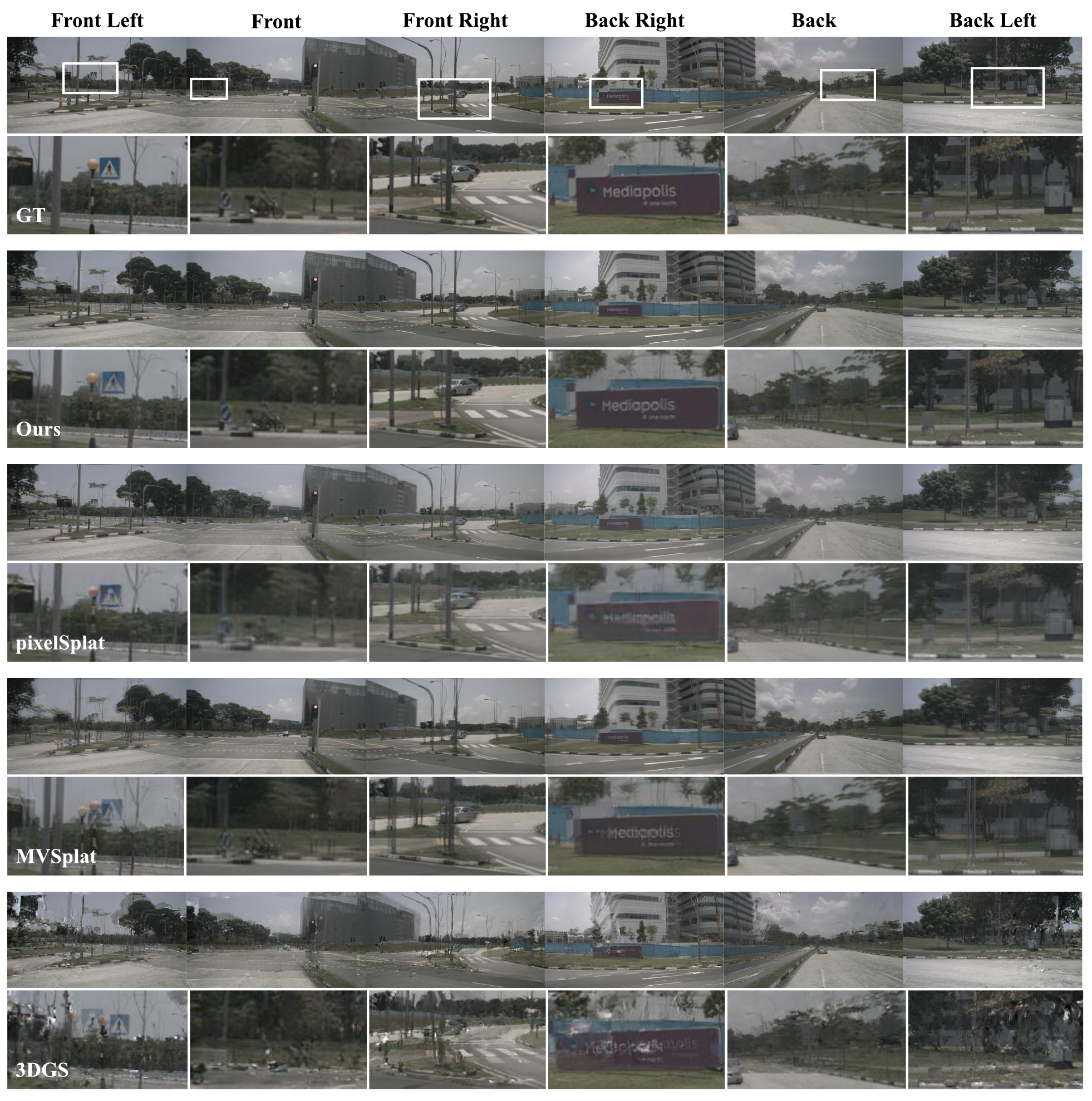}
    \caption{Complete visualization results in the main paper. Compared with the state-of-the-art feed-forward and scene-optimized reconstruction methods, our method reduces artifacts and produces more detailed surround-view scenes.}
    \label{fig: supp_full_vis}
\end{figure*}

\begin{figure*}[!t]
    \centering
    \includegraphics[width=0.95\textwidth]{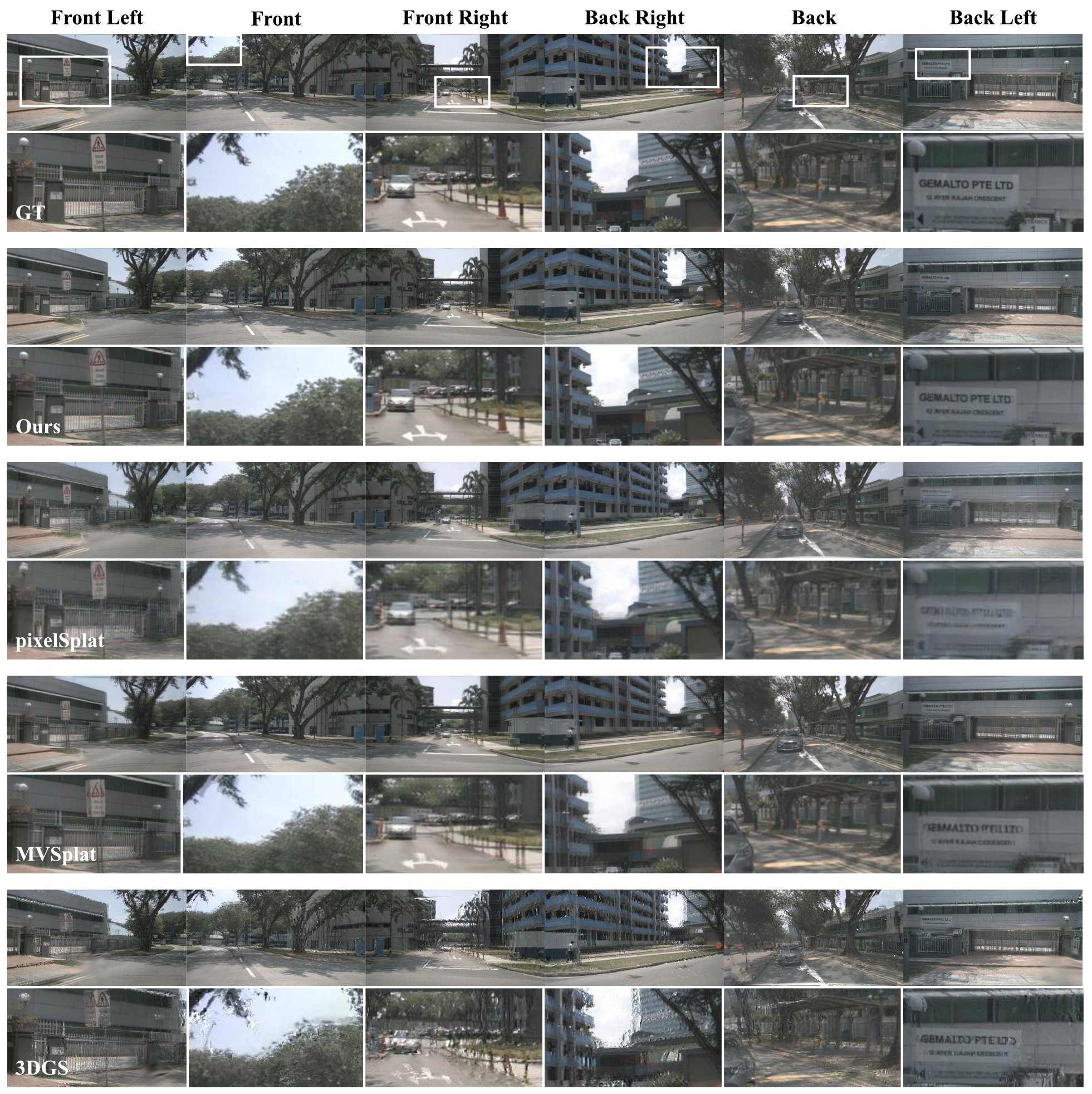}
    \caption{More visualization results on the nuScenes dataset.}
    \label{fig: supp_more_vis}
\end{figure*}

We present novel view visualization results comparing our method with pixelSplat, MVSplat, and 3DGS in MF mode, including the complete visualization results in the main paper (Figure~\ref{fig: supp_full_vis}) and additional 
visualization results (Figure~\ref{fig: supp_more_vis}).\par

\end{document}